\documentclass[11pt,a4paper]{article}
\usepackage{balance}
\usepackage{booktabs}
\usepackage[hyperref]{mrqa}
\usepackage{times}
\usepackage{latexsym}
\usepackage{multirow}
\usepackage[utf8]{inputenc}
\usepackage{graphicx}
\usepackage{amsmath}
\usepackage{makecell}

\usepackage{url}

\aclfinalcopy 

\newcommand*\samethanks[1][\value{footnote}]{\footnotemark[#1]}

\title{An Exploration of Data Augmentation and Sampling Techniques for Domain-Agnostic Question Answering}

\author{Shayne Longpre\thanks{\hspace{2mm}equal contribution}, Yi Lu\samethanks, Zhucheng 
Tu\samethanks, Chris DuBois\\
Apple Inc.\\
{\tt \small \{slongpre, ylu7, zhucheng\_tu, cdubois\}@apple.com }}

\date{}

\begin{document}
\maketitle
\begin{abstract}
To produce a domain-agnostic question answering model for the Machine Reading Question Answering (MRQA) 2019 Shared Task, we investigate the relative benefits of large pre-trained language models, various data sampling strategies, as well as query and context paraphrases generated by back-translation. 
We find a simple negative sampling technique to be particularly effective, even though it is typically used for datasets that include unanswerable questions, such as SQuAD 2.0. 
When applied in conjunction with per-domain sampling, our XLNet \cite{yang2019xlnet}-based submission achieved the second best Exact Match and F1 in the MRQA leaderboard competition.
\end{abstract}

\section{Introduction}

Recent work has demonstrated that generalization remains a salient challenge in extractive question 
answering \cite{talmor2019multiqa,yogatama2019learning}. It is especially difficult to generalize
to a target domain without similar training data, or worse, any knowledge of the domain’s distribution.
This is the case for the MRQA Shared Task.\footnote{\url{https://mrqa.github.io/shared}} Together, 
these two factors demand a representation that generalizes broadly, and rules out the usual
assumption that more data in the training domain will necessarily improve performance on the target 
domain.
Consequently, we adopt the overall approach of curating our input data and learning regime to 
encourage representations that are not biased by any one domain or distribution.

 As a requisite first step to a representation that generalizes, transfer learning (in the form of large 
 pre-trained language models such as 
 \citet{peters2018deep,howard2018universal,devlin2019bert,yang2019xlnet}), offers a solid 
foundation. We compare BERT and XLNet, leveraging Transformer based models \cite{vaswani2017attention} 
pre-trained on significant quantities of unlabelled text. Secondly, we 
identify how the domains of our training data correlate with the performance of ``out-domain" development sets.
This serves as a proxy for the impact these different sets may have on a held-out test set, 
as well as evidence of a representation that generalizes. Next we explore data 
sampling and augmentation strategies to better leverage our 
available supervised data. 

To our surprise, the more sophisticated techniques including back-translated augmentations (even sampled with active learning strategies) yield no noticeable improvement. In contrast, much simpler techniques offer significant improvements. In particular, negative samples designed to teach the model when to abstain from predictions prove highly effective out-domain. We hope our analysis and results, both positive and negative, inform the challenge of 
generalization in multi-domain question answering.

We begin with an overview of the data and techniques used in our system, before discussing 
experiments and results.

\section{Data}
\label{sec:data}

We provide select details of the MRQA data as they pertain to our sampling strategies delineated later. For greater detail refer to the MRQA task description. 

Our training data 
consists of six separately collected QA datasets. We refer to these and their associated development 
sets as ``in-domain'' (ID). We are also provided with six ``out-domain'' (OD) development sets 
sourced from other QA 
datasets. In Table~\ref{table:data} we tabulate the number of 
``examples'' (question-context pairs), ``segments'' (the question combined with a portion of the context), and ``no-answer'' 
(NA) segments (those without a valid answer span).

\begin{table}[t]
	\centering
	\resizebox{\columnwidth}{!}{
			\begin{tabular}{lrrr}
				\toprule
				Dataset & Examples & Segments & NA (\%) \\
				\midrule
				SQuAD \cite{rajpurkar2016squad} & 87K & 87K & 0.1 \\
				SearchQA \cite{dunn2017searchqa} & 117K & 657K & 56.3 \\
				NaturalQuestions \cite{Kwiatkowski} & 104K & 189K & 36.3 \\
				TriviaQA \cite{joshi2017triviaqa} & 62K & 337K & 57.3 \\
				HotpotQA \cite{yang2018hotpotqa} & 73K & 73K & 0.3 \\
				NewsQA \cite{trischler2017newsqa} & 74K & 214K & 49.0 \\
				\midrule
				\textbf{Total} & \textbf{517K} & \textbf{1557K} &\textbf{47.3} \\
				\bottomrule
			\end{tabular}}
	\caption{Number of examples (question-context pairs), segments (question-context chunks), and 
	the percentage of
	No Answer (NA) segments within each dataset.}
	\label{table:data}
\end{table}

To clarify these definitions, consider examples with long context sequences. We found it necessary to 
break these examples' contexts into multiple segments in order to satisfy computational memory 
constraints. Each of these segments may or may not contain the gold answer span. A segment without 
an answer span we term ``no-answer''. To illustrate this pre-processing, consider question, context
pair $(q,c)$ where we impose a maximum sequence length of $M$ tokens. If
$len(c)>M$ then we create multiple overlapping input segments $(q,c_{1})$,
$(q,c_{2})$, $...$, $(q,c_{k})$ where each $c_{i}$ contains only a portion
of the larger context $c$. The sliding window that generates these
chunks is parameterized by the document stride $D$, and the maximum
sequence length $M$, shown below in Equation~\ref{eq:segments}.
\vspace{-2mm}

\begin{equation}
\label{eq:segments}
(q,c)\rightarrow \big\{ (q, c_{i \cdot D:M+i \cdot D}), \forall i \in [0,k] \big\}
\end{equation}

The frequencies presented in Table~\ref{table:data} are based on our settings of $M=512$ and 
$D=128$.

\section{System Overview}
\label{sec:modeling}

\subsection{XLNet}

While we used BERT Base \cite{devlin2019bert} for most of our experimentation, we
used XLNet Large \cite{yang2019xlnet} for our final 
submission. At the time of submission
this model held state-of-the-art results on several NLP benchmarks
including GLUE \cite{wang2018glue}. Leveraging the Transformer-XL architecture 
\cite{dai2019transformer}, a ``generalized autoregressive pretraining'' method, and
much more training data than BERT, its representation provided a strong
source of transfer learning. In keeping with XLNet's question answering module, we also computed the 
end logits based 
on the ground truth of the start position during training time, and used beam search over
the end logits at inference time. We based our code on the HuggingFace 
implementation of BERT and XLNet, and used the pre-trained models in the GitHub repository.\footnote{\url{https://github.com/huggingface/pytorch-transformers}}
Our implementation modifies elements of the tokenization, modeling, and training procedure. Specifically,
we remove whitespace tokenization and other pre-processing features that are not necessary for MRQA-tokenized
data. We also add subepoch checkpoints and validation, per dataset sampling, and improved
post-processing to select predicted text without special tokens or unusual spacing.

\subsection{Domain Sampling}
\label{sec:domain}

For the problem of generalizing to an unseen and out-domain test set,
it's important not to overfit to the training distribution. Given the selection of diverse training sources, 
domains,
and distributions within MRQA we posed the following questions. Are all training
sources useful to the target domains? Will multi-domain training
partially mitigate overfitting to any given training set? Is it always
appropriate to sample equally from each? 

To answer these questions, we fine-tuned a variety of specialized models on the BERT Base Cased 
(BBC)
pre-trained model. Six models were each fine-tuned once on their respective in-domain training set.
A multi-domain model was trained on the union of these six in-domain training sets. Lastly, we used
this multi-domain model as the starting point for fine-tuning six more models, one for each in-domain training set.
In total we produced six dataset-specialized models each fine-tuned once, one multi-domain model, and six 
dataset-specialized models each fine-tuned twice.

\begin{figure*}[htp]
    \centering
    \includegraphics[width=1\linewidth]{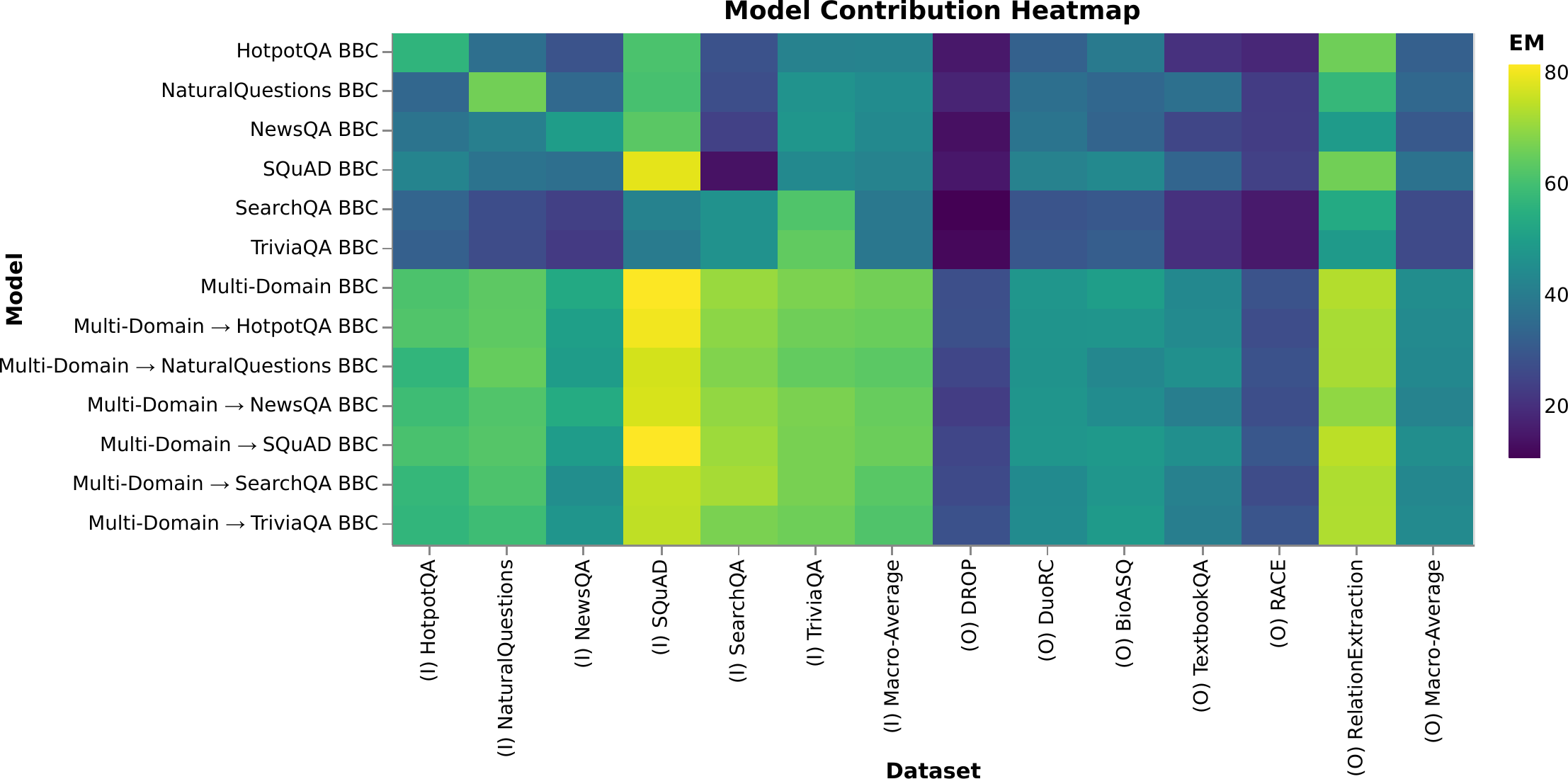}
    \caption{Heatmap of Exact Match (EM) for BERT Base Cased 
    (BBC) models, the top six fine-tuned directly on each training dataset, and the bottom six 
    fine-tuned on multi-domain before being fine-tuned on each training dataset.}
    \label{fig:heatmap}
\end{figure*}

There are a few evident trends. The set of models which were
first fine-tuned on the multi-domain dataset achieved
higher Exact Match (EM) almost universally than those which weren't. This improvement 
extends not just to in-domain datasets, but also to out-domain development sets. In 
Figure~\ref{fig:heatmap} we observe these models 
on the Y-axis, and their Exact Match (EM) scores on each in-domain and out-domain development set. 
This confirms
the observations from \citet{talmor2019multiqa} that multi-domain training improves robustness and 
generalization broadly, and suggests that a variety
of question answering domains is significant across domains. Interestingly, the second round of 
fine-tuning,
this time on a specific domain, did not cause models to significantly, or catastrophically forget what 
they learned in the initial, multi-domain fine-tuning. This is clear from comparing the generic 
``Multi-Domain BBC'' to those models fine-tuned on top of it, such as ``Multi-Domain $\rightarrow$ 
SQuAD FT BBC''.

Secondly, we observe that the models we fine-tune on SearchQA \cite{dunn2017searchqa} and 
TriviaQA \cite{joshi2017triviaqa} achieve relatively poor results across all sets (in-domain and 
out-domain) aside from themselves.
The latter datasets are both Jeopardy-sourced,
distantly supervised, long context datasets. In contrast, the SQuAD \cite{rajpurkar2016squad} 
fine-tuned model achieves the best results on
both in and out-domain ``Macro-Average'' Exact Match. Of the models with multi-domain 
pre-fine-tuning NewsQA, SearchQA, and TriviaQA performed the worst on the out-domain (O) 
Macro-Average. As such we modified our sampling distribution to avoid oversampling them and risk 
degrading generalization performance. This risk is particularly prevalent for SearchQA, the largest 
dataset by number of examples. Additionally, its long contexts generate 657K segments, double that of 
the next largest dataset (Table~\ref{table:data}). This was exacerbated further when we initially included 
the nearly 10 occurrences of each
detected answer. TriviaQA shares this characteristic, though not quite as drastically. Accordingly,
for our later experiments we chose not to use all instances of a detected
answer, as this would further skew our multi-domain samples towards
SearchQA and TriviaQA, and increase the number of times contexts from these sets are repeated as 
segments. We also chose,
for many experiments, to sample fewer examples of SearchQA
than our other datasets, and found this to improve F1 marginally across configurations.

\subsection{Negative Sampling}

\begin{table}[t]
	\centering
	\resizebox{\columnwidth}{!}{
		\begin{small}
			\begin{tabular}{lcccccc}
				\toprule
				&  & & \multicolumn{2}{c}{In-Domain} & \multicolumn{2}{c}{Out-Domain} \\
				NA & Model & MSL & EM & F1 & EM & F1 \\
				\midrule
				No & BBC & 200 & 65.70 & 75.98 & 45.80 & 56.78 \\
				& BBC & 512 & 65.29 & 76.01 & 45.59 & 57.40 \\
				& XBC & 200 & 43.78 & 65.24 & 43.78 & 52.12 \\
				& XBC & 512 & 65.91 & 74.93 & 49.59 & 59.61 \\
				\midrule
				Yes & BBC & 200 & 66.11 & 76.41 & 46.19 & 57.51 \\
				& BBC & 512 & 66.20 & 76.77 & 46.28 & 58.00 \\
				& XBC & 200 & 68.67 & 77.69 & 50.04 & 59.68 \\
				& XBC & 512 & \textbf{70.04} & \textbf{79.15} & \textbf{50.71} & \textbf{61.16} \\
				\bottomrule
			\end{tabular}
	\end{small}}
	\caption{Model performance including or excluding No-Answer (NA) segments in training. We 
	examine how these results vary with the max sequence length (MSL). 
		BBC refers to BERT Base Cased and XBC refers to XLNet Base Cased.}
	\label{table:NA-stats}

\end{table}

While recent datasets such as SQuAD 2.0 \cite{rajpurkar2018know} and Natural Questions
\cite{Kwiatkowski} have extended extractive question answering to include a No Answer
option, in the traditional formulation of the problem there is no notion of a
negative class. Formulated as such, the MRQA Shared Task guarantees
the presence of an answer span within each example. However, this is not guaranteed within each 
segment, producing NA segments.

At inference time we compute the most probable answer span for
each segment separately and then select the best span across all segments
of that ($\bf{q}$, $\bf{c}$) example to be the one with the highest probability. This is computed as the 
sum of the start and end span probabilities. At training time, typically the NA segments are discarded 
altogether. However, this causes
a discrepancy between train and inference time, as ``Negative''
segments are only observed in the latter. 

To address this, we include naturally occurring ``Negative'' segments, and add an
abstention option for the model. For each Negative segment, we set the indices
for both the start and end span labels to point to the \texttt{[CLS]} token. This
gives our model the option to abstain from selecting a span in a given
segment. Lastly, at inference time we select the highest probability answer across all segments,
excluding the No Answer \texttt{[CLS]} option.

Given that 47.3\% of all input segments are NA, as shown in Table~\ref{table:data}, its unsurprising 
their inclusion significantly impacted training time and results. We find that this simple form of 
Negative Sampling yields non-trivial improvements on MRQA (see Table~\ref{table:NA-stats}). 
We hypothesize
this is primarily because a vaguely relevant span of tokens amid a completely irrelevant NA segment 
would monopolize the predicted probabilities. Meanwhile the actual answer span likely appears in a 
segment that may contain many competing spans of relevant text, each attracting some probability 
mass. As we would expect, the improvement this technique offers is magnified where the context is 
much
longer than $M$. To our knowledge this technique is still not prevalent in purely extractive question 
answering,
though \citet{alberti2019bert} cite it as a key contributor to their strong baseline
on Google's Natural Questions.

\subsection{Paraphrasing by Back-Translation}

\citet{yu2018qanet} showed that generating context paraphrases via back-translation provides 
significant improvements for reading comprehension on the competitive SQuAD 1.1 benchmark. We 
emulate this approach to add further quantity and variety to our data distribution, with the hope that it 
would produce similarly strong results for out-domain generalization. To extend their work, we 
experiment with both query and context paraphrases generated by back-translation. Leveraging the 
same open-sourced TensorFlow NMT codebase,\footnote{\url{https://github.com/tensorflow/nmt}} we 
train an 8-layer \texttt{seq2seq} model with attention on the WMT16 News English-German task, 
obtaining a BLEU score of 28.0 for translating from English to German and 25.7 for German to English, 
when evaluated on the \texttt{newstest2015} dataset. We selected German as our back-translation 
language due to ease of reproducibility, given the public benchmarks published in the \texttt{nmt} 
repository.

For generating query paraphrases, we directly feed each query into the NMT model after performing 
tokenization and byte pair encoding. For generating context paraphrases, we first use 
SpaCy to segment each context into sentences,\footnote{\url{https://spacy.io/}} using the 
\texttt{en\_core\_web\_sm} model. Then, we translate each sentence independently, following the same 
procedure as we do for each query. In the course of generating paraphrases, we find decoded 
sequences are occasionally empty for a given context or query input. For these cases we keep the 
original sentence.

We attempt to retrieve the new answer span using string matching, and where that fails we employed 
the the same heuristic described in \citet{yu2018qanet} to obtain a new, estimated answer. Specifically, 
this involves finding the character-level 2-gram overlap of every token in the paraphrase sentence with 
the start and end token of the original answer. The score is computed as the Jaccard similarity 
between the sets of character-level 2-grams in the original answer token and new sentence token. The 
span of text between the two tokens that has the highest combined score, passing a minimum 
threshold, is selected as the new answer. In cases where there is no score above the threshold, no 
answer is generated. Any question in each context without an answer is omitted, and any paraphrased 
example without at least one question-answer pair is discarded.

\subsubsection{Augmentation Strategy}

For every query and context pair $(q, c)$, we used our back-translation model to generate a query paraphrase $q'$ and a context paraphrase $c'$. 
We then create a new pair that includes the paraphrase $q'$ instead of $q$ with probability $P_{q}(x)$, and independently we choose the paraphrase $c'$ over $c$ with probability $P_{c}(x)$. 
If either $q'$ or $c'$ is sampled, we add this augmented example to the training data. 
This sampling strategy allowed us flexibility in how often we include query or context augmentations. 

\subsubsection{Active Learning}
\label{sec:active}

Another method of sampling our data augmentations was motivated by principles in 
active learning \cite{settles2009active}. Rather than sampling uniformly, might we prioritize the 
more challenging examples for augmentation? This is motivated by the idea that many augmentations may
not be radically different from the original data points, and may consequently carry less useful, repetitive 
signals.

To quantify the difficulty of an example we used $1-F1$ score computed for our best model. 
We chose F1 as it provides a continuous rather than binary value, and is robust to a model that may 
select the wrong span, but contains the correct 
answer text. Other metrics, such as loss or Exact Match do not provide both these benefits.

For each example we derived its probability weighting from its F1 score. This weight replaces the 
uniform probability previously used to draw samples for query and context augmentations. We devised 
three weighting strategies, to experiment with different distributions. We refer to these as the hard, 
moderate and soft distributions. Each distribution employs its own scoring function $S_x$ 
(Equation~\ref{eq:active-learning1}), which is normalized across all examples to 
determine the probability of drawing that sample (Equation~\ref{eq:active-learning2}).
\vspace{-2mm}

\begin{equation}
\label{eq:active-learning1}
S(x) = \begin{cases}
1 - F1(x) + \epsilon &\text{Hard Score} \\
2 - F1(x) &\text{Moderate Score} \\
3 - F1(x) &\text{Soft Score} \\
\end{cases}
\end{equation}

\begin{equation}
\label{eq:active-learning2}
P(x) = \frac{S(x)}{\Sigma_{i = 1..n} S(i) }
\end{equation}

The hard scoring function allocates negligible probability to examples with $F1=1$, emphasizing the 
hardest examples the most of the three distributions. We used an $\epsilon$  value of 0.01 to prevent 
any example from having a zero sample probability. The moderate and soft scoring functions penalize 
correct predictions less aggressively, smoothing the distribution closer to uniform.

\section{Experiments and Discussion}

\begin{table*}[t]
	\centering
	\resizebox{\linewidth}{!}{
			\begin{tabular}{lllccccccc|ccccccc}
				\toprule
				&  & & \multicolumn{7}{c}{In-Domain F1} & \multicolumn{7}{c}{Out-Domain F1} \\
				Mode & $P_q(x)$ & $P_c(x)$ & HotpotQA & \thead{Natural\\Questions} & NewsQA & SearchQA & SQuAD & TriviaQA & Macro-Average & BioASQ & DROP & DuoRC & RACE & \thead{Relation\\Extraction} & TextbookQA & Macro-Average \\
				\midrule
				-- & 0 & 0 & 82.62 & 82.15 & 72.52 & 82.80 & \textbf{94.50} & 78.28 & 82.14 & \textbf{73.00} & 63.52 & \textbf{65.68} & \textbf{53.25} & 88.49 & 64.38 & \textbf{68.07} \\ 
				\midrule
				R & 0.2 & 0.2 & 82.42 & 82.29 & 72.45 & 83.20 & 94.09 & \textbf{79.44} & 82.32 & 70.45 & 63.97 & 62.75 & 52.66 & 88.09 & 63.28 & 66.87 \\ 
				& 0.2 & 0.4 & 82.59 & \textbf{82.51} & 72.30 & 84.50 & 94.35 & 79.09 & \textbf{82.56} & 72.02 & \textbf{64.29} & 63.61 & 52.32 & \textbf{88.85} & 64.12 & 67.54 \\ 
				& 0.4 & 0.4 & 82.58 & 82.28 & 71.72 & 83.80 & 94.02 & 77.78 & 82.03 & 69.60 & 63.45 & 63.56 & 52.74 & 88.22 & 63.67 & 66.87 \\ 
				\midrule
				S & 0.2 & 0.2 & 82.44 & 82.10 & 72.06 & 83.67 & 94.32 & 76.58 & 81.86 & 70.47 & 64.14 & 63.15 & 52.61 & 88.37 & 63.60 & 67.06 \\ 
				& 0.2 & 0.4 & 82.50 & 81.69 & 72.43 & 84.46 & 93.98 & 76.80 & 81.98 & 70.79 & 60.62 & 63.48 & 52.38 & 87.38 & 62.07 & 66.12 \\ 
				& 0.4 & 0.4 & 82.07 & 82.15 & 72.07 & 84.20 & 93.99 & 77.20 & 81.95 & 71.34 & 62.64 & 62.81 & 50.65 & 87.60 & 63.12 & 66.36 \\ 
				\midrule
				M & 0.2 & 0.2 & \textbf{82.72} & 82.26 & 72.22 & 83.45 & 94.12 & 76.55 & 81.89 & 71.46 & 63.89 & 63.29 & 51.67 & 87.98 & \textbf{64.85} & 67.19 \\ 
				& 0.2 & 0.4 & 82.41 & 82.15 & \textbf{72.60} & \textbf{84.88} & 93.85 & 77.34 & 82.20 & 71.66 & 63.89 & 62.12 & 52.67 & 88.03 & 64.05 & 67.07 \\ 
				& 0.4 & 0.4 & 82.55 & 82.09 & 72.57 & 84.30 & 94.19 & 76.97 & 82.11 & 71.13 & 63.03 & 62.58 & 51.65 & 87.76 & 64.67 & 66.80 \\ 
				\midrule
				H & 0.2 & 0.2 & 81.68 & 81.15 & 70.55 & 80.51 & 94.05 & 74.80 & 80.46 & 70.60 & 62.55 & 61.96 & 52.23 & 87.87 & 61.16 & 66.06 \\ 
				& 0.2 & 0.4 & 82.05 & 81.45 & 70.84 & 81.92 & 94.18 & 75.49 & 80.99 & 72.89 & 62.29 & 63.30 & 51.66 & 87.63 & 62.00 & 66.63 \\ 
				& 0.4 & 0.4 & 81.93 & 81.45 & 71.67 & 81.71 & 93.92 & 75.96 & 81.11 & 71.26 & 61.52 & 62.06 & 51.36 & 86.91 & 60.18 & 65.55 \\ 
				\bottomrule
			\end{tabular}}
	\caption{F1 scores for data augmentation using different proportions of query and context 
		paraphrasing and different sampling distributions on XLNet Large Cased, on individual datasets. R, S, M, H refer to random, soft, moderate, and hard modes from Section~\ref{sec:active} respectively.}
	\label{table:active-learning-dataset}
\end{table*}

\begin{table}[t]
	\centering
	\resizebox{\columnwidth}{!}{
		\begin{small}
		\begin{tabular}{lcc}
			\toprule
			& \multicolumn{2}{c}{Out-Domain} \\
			Dataset & EM & F1 \\
			\midrule
			BioASQ \cite{tsatsaronis2015overview} & 60.28 & 71.98 \\
			DROP \cite{dua2019drop} & 48.50 & 58.90 \\
			DuoRC \cite{saha2018duorc} & 53.29 & 63.36 \\
			RACE \cite{lai2017race} & 39.35 & 53.87 \\
			RelationExtraction \cite{levy2017zero} & 79.20 & 87.85 \\
			TextbookQA \cite{kembhavi2017you} & 56.50 & 65.54 \\ 
			\midrule
			Macro-Average & 56.19 & 66.92 \\
			\bottomrule
		\end{tabular}
		\end{small}}
	\caption{Breakdown of hidden development set results by dataset using our best XLNet Large model.}
	\label{table:dev-results}
\end{table}

\begin{table}[t]
	\centering
	\resizebox{\columnwidth}{!}{
			\begin{tabular}{lcc}
				\toprule
				Submission & EM & F1 \\
				\midrule
				D-NET (Baidu) & \textbf{60.39} & \textbf{72.55} \\
				Ours (Apple) & 59.47 & 70.75 \\
				FT\_XLNet (HIT) & 58.37 & 70.54 \\
				HLTC (HKUST) & 56.59 & 68.98 \\
				BERT-cased-whole-word (Aristo@AI2) & 53.52 & 66.27 \\
				XERO (Fuji Xerox) & 52.41 & 66.11 \\
				BERT-large + Adv. Training (Team 42-alpha) & 48.91 & 62.19 \\
				\midrule
				\textit{BERT large baseline} (MRQA Organizers) & 48.20 & 61.76 \\
				 \textit{BERT base baseline} (MRQA Organizers) & 45.54 & 58.50 \\
				\bottomrule
			\end{tabular}}
	\caption{Macro-Average EM and F1 on the held-out leaderboard test sets.}
	\label{table:competition-results}
\end{table}

During our experimentation process we used our smallest model BERT Base Cased (BBC) for the most 
expensive sampling explorations (Figure~\ref{fig:heatmap}), XLNet Base Cased (XBC) to confirm our 
findings extended to XLNet (Table~\ref{table:NA-stats}), and XLNet Large Cased (XLC) as the initial 
basis for our final submission contenders (Table~\ref{table:active-learning-dataset}). 

Our training procedure for each model involved fine-tuning the Transformer over two epochs, each 
with three validation checkpoints. The checkpoint with the highest Out-Domain Macro-Average 
(estimated from a $2,000$ dev-set subsample) was selected as the best for that training run. Our 
multi-domain dataset originally consisted of 75k examples from every training set, and using every 
detected answer. We modified this to a maximum of 120k samples from each dataset, 100k from 
SearchQA, and using only one detected answer per example; given our findings in 
Section~\ref{sec:domain}.

We trained every model on $8$ NVIDIA Tesla V100 GPUs.
For BBC and XBC we used a learning rate of $5e-5$, single-GPU batch size of $25$, and gradient 
accumulation of $1$, yielding an effective batch size of $200$. For XLC we used a learning rate of 
$2e-5$, single-GPU batch size of $6$, and gradient accumulation of $3$, yielding an effective batch 
size of $6 \cdot 8 \cdot 3=144$. We found the gradient accumulation and lower learning rate critical to achieve 
training stability.

We conduct several experiments to evaluate the various sampling and augmentation strategies 
discussed in Section~\ref{sec:modeling}. In Table~\ref{table:NA-stats} we examine the impact of 
including No Answer segments in our training set. We found this drastically out-performed the typical 
practice of excluding these segments. This effect was particularly noticeable on datasets with 
longer sequences. As expected, the improvement is exaggerated at the shorter max sequence length 
(MSL) of 200, where including NA segments increases Out-Domain EM from $43.78$ to $ 50.04$ on 
the XBC model.

Next, we evaluate our back-translated query and context augmentations using the sampling strategies 
described in Section~\ref{sec:active}. To select the best  $P_q(x)$, $P_c(x)$ and sampling strategy we 
conducted the following search. First we explored sampling probabilities $0.2$, $0.4$, $0.6$, $0.8$, 
$1.0$ for query and context separately, using random sampling, and subsequently we combined them 
using  values informed from the previous exploration, this time searching over sampling strategies: 
random, soft, moderate and hard. We present the best results in Table~\ref{table:active-learning-dataset} and 
conclude that these data augmentations did not help in-domain or out-domain performance.
While we observed small boosts to metrics on BBC using this technique, no such gains were found on XLC.
We suspect this is because (a) large pre-trained language models such as XLC already capture the linguistic variations in language introduced by paraphrased examples quite well,
and (b) we already have a plethora of diverse training data from the distributions these augmentations are derived from.
It is not clear if the boosts QANet \citet{yu2018qanet} observed on SQuAD 1.1 would still apply with the additional diversity provided by the five additional QA datasets for training.
We notice that SearchQA and TriviaQA benefit the most from some form of data augmentation, both by more than one F1 point.
Both of these are distantly supervised, and have relatively long contexts.

Our final submission leverages our fine-tuned XLC configuration, with domain and negative sampling. We 
omit the data augmentation and active sampling techniques which we did not find to aid 
out-domain performance.
The results of the leaderboard Out-Domain Development set and final test set results are shown in Table ~\ref{table:dev-results} and Table~\ref{table:competition-results} respectively.

\section{Conclusion}

This paper describes experiments on various competitive pre-trained models (BERT, XLNet), 
domain sampling strategies, negative sampling, data augmentation via back-translation, and active 
learning. We determine which of these strategies help and hurt multi-domain generalization, finding 
ultimately that some of the simplest techniques offer surprising improvements. The most significant
benefits came from sampling No Answer segments, which proved to be particularly important for
training extractive models on long sequences. In combination these findings culminated in the
second ranked submission on the MRQA-19 Shared Task.

\bibliographystyle{acl_natbib}
\bibliography{mrqa}

\begin{thebibliography}{25}
\expandafter\ifx\csname natexlab\endcsname\relax\def\natexlab#1{#1}\fi

\bibitem[{Alberti et~al.(2019)Alberti, Lee, and Collins}]{alberti2019bert}
Chris Alberti, Kenton Lee, and Michael Collins. 2019.
\newblock A {BERT} baseline for the {Natural Questions}.
\newblock \emph{arXiv preprint arXiv:1901.08634}.

\bibitem[{Dai et~al.(2019)Dai, Yang, Yang, Cohen, Carbonell, Le, and
  Salakhutdinov}]{dai2019transformer}
Zihang Dai, Zhilin Yang, Yiming Yang, William~W Cohen, Jaime Carbonell, Quoc~V
  Le, and Ruslan Salakhutdinov. 2019.
\newblock {Transformer-XL}: Attentive language models beyond a fixed-length
  context.
\newblock \emph{arXiv preprint arXiv:1901.02860}.

\bibitem[{Devlin et~al.(2019)Devlin, Chang, Lee, and
  Toutanova}]{devlin2019bert}
Jacob Devlin, Ming-Wei Chang, Kenton Lee, and Kristina Toutanova. 2019.
\newblock {BERT}: Pre-training of deep bidirectional transformers for language
  understanding.
\newblock In \emph{Proceedings of the 2019 Conference of the North American
  Chapter of the Association for Computational Linguistics: Human Language
  Technologies, Volume 1 (Long and Short Papers)}, pages 4171--4186.

\bibitem[{Dua et~al.(2019)Dua, Wang, Dasigi, Stanovsky, Singh, and
  Gardner}]{dua2019drop}
Dheeru Dua, Yizhong Wang, Pradeep Dasigi, Gabriel Stanovsky, Sameer Singh, and
  Matt Gardner. 2019.
\newblock {DROP}: A reading comprehension benchmark requiring discrete
  reasoning over paragraphs.
\newblock In \emph{Proceedings of the 2019 Conference of the North American
  Chapter of the Association for Computational Linguistics: Human Language
  Technologies, Volume 1 (Long and Short Papers)}, pages 2368--2378.

\bibitem[{Dunn et~al.(2017)Dunn, Sagun, Higgins, Guney, Cirik, and
  Cho}]{dunn2017searchqa}
Matthew Dunn, Levent Sagun, Mike Higgins, V~Ugur Guney, Volkan Cirik, and
  Kyunghyun Cho. 2017.
\newblock {SearchQA}: A new {Q}\&{A} dataset augmented with context from a
  search engine.
\newblock \emph{arXiv preprint arXiv:1704.05179}.

\bibitem[{Howard and Ruder(2018)}]{howard2018universal}
Jeremy Howard and Sebastian Ruder. 2018.
\newblock Universal language model fine-tuning for text classification.
\newblock In \emph{Proceedings of the 56th Annual Meeting of the Association
  for Computational Linguistics (Volume 1: Long Papers)}, pages 328--339.

\bibitem[{Joshi et~al.(2017)Joshi, Choi, Weld, and
  Zettlemoyer}]{joshi2017triviaqa}
Mandar Joshi, Eunsol Choi, Daniel Weld, and Luke Zettlemoyer. 2017.
\newblock {TriviaQA}: A large scale distantly supervised challenge dataset for
  reading comprehension.
\newblock In \emph{Proceedings of the 55th Annual Meeting of the Association
  for Computational Linguistics (Volume 1: Long Papers)}, pages 1601--1611.

\bibitem[{Kembhavi et~al.(2017)Kembhavi, Seo, Schwenk, Choi, Farhadi, and
  Hajishirzi}]{kembhavi2017you}
Aniruddha Kembhavi, Minjoon Seo, Dustin Schwenk, Jonghyun Choi, Ali Farhadi,
  and Hannaneh Hajishirzi. 2017.
\newblock Are you smarter than a sixth grader? textbook question answering for
  multimodal machine comprehension.
\newblock In \emph{2017 IEEE Conference on Computer Vision and Pattern
  Recognition (CVPR)}, pages 5376--5384. IEEE.

\bibitem[{Kwiatkowski et~al.(2019)Kwiatkowski, Palomaki, Redfield, Collins,
  Parikh, Alberti, Epstein, Polosukhin, Kelcey, Devlin, Lee, Toutanova, Jones,
  Chang, Dai, Uszkoreit, Le, and Petrov}]{Kwiatkowski}
Tom Kwiatkowski, Jennimaria Palomaki, Olivia Redfield, Michael Collins, Ankur
  Parikh, Chris Alberti, Danielle Epstein, Illia Polosukhin, Matthew Kelcey,
  Jacob Devlin, Kenton Lee, Kristina~N. Toutanova, Llion Jones, Ming-Wei Chang,
  Andrew Dai, Jakob Uszkoreit, Quoc Le, and Slav Petrov. 2019.
\newblock \href
  {https://tomkwiat.users.x20web.corp.google.com/papers/natural-questions/main-1455-kwiatkowski.pdf}
  {Natural questions: a benchmark for question answering research}.
\newblock \emph{Transactions of the Association of Computational Linguistics}.

\bibitem[{Lai et~al.(2017)Lai, Xie, Liu, Yang, and Hovy}]{lai2017race}
Guokun Lai, Qizhe Xie, Hanxiao Liu, Yiming Yang, and Eduard Hovy. 2017.
\newblock {RACE}: Large-scale reading comprehension dataset from examinations.
\newblock In \emph{Proceedings of the 2017 Conference on Empirical Methods in
  Natural Language Processing}, pages 785--794.

\bibitem[{Levy et~al.(2017)Levy, Seo, Choi, and Zettlemoyer}]{levy2017zero}
Omer Levy, Minjoon Seo, Eunsol Choi, and Luke Zettlemoyer. 2017.
\newblock Zero-shot relation extraction via reading comprehension.
\newblock In \emph{Proceedings of the 21st Conference on Computational Natural
  Language Learning (CoNLL 2017)}, pages 333--342.

\bibitem[{Peters et~al.(2018)Peters, Neumann, Iyyer, Gardner, Clark, Lee, and
  Zettlemoyer}]{peters2018deep}
Matthew Peters, Mark Neumann, Mohit Iyyer, Matt Gardner, Christopher Clark,
  Kenton Lee, and Luke Zettlemoyer. 2018.
\newblock Deep contextualized word representations.
\newblock In \emph{Proceedings of the 2018 Conference of the North American
  Chapter of the Association for Computational Linguistics: Human Language
  Technologies, Volume 1 (Long Papers)}, pages 2227--2237.

\bibitem[{Rajpurkar et~al.(2018)Rajpurkar, Jia, and Liang}]{rajpurkar2018know}
Pranav Rajpurkar, Robin Jia, and Percy Liang. 2018.
\newblock Know what you don’t know: Unanswerable questions for {SQuAD}.
\newblock In \emph{Proceedings of the 56th Annual Meeting of the Association
  for Computational Linguistics (Volume 2: Short Papers)}, pages 784--789.

\bibitem[{Rajpurkar et~al.(2016)Rajpurkar, Zhang, Lopyrev, and
  Liang}]{rajpurkar2016squad}
Pranav Rajpurkar, Jian Zhang, Konstantin Lopyrev, and Percy Liang. 2016.
\newblock {SQuAD}: 100,000+ questions for machine comprehension of text.
\newblock In \emph{Proceedings of the 2016 Conference on Empirical Methods in
  Natural Language Processing}, pages 2383--2392.

\bibitem[{Saha et~al.(2018)Saha, Aralikatte, Khapra, and
  Sankaranarayanan}]{saha2018duorc}
Amrita Saha, Rahul Aralikatte, Mitesh~M Khapra, and Karthik Sankaranarayanan.
  2018.
\newblock {DuoRC}: Towards complex language understanding with paraphrased
  reading comprehension.
\newblock In \emph{Proceedings of the 56th Annual Meeting of the Association
  for Computational Linguistics (Volume 1: Long Papers)}, pages 1683--1693.

\bibitem[{Settles(2009)}]{settles2009active}
Burr Settles. 2009.
\newblock Active learning literature survey.
\newblock Technical report, University of Wisconsin-Madison Department of
  Computer Sciences.

\bibitem[{Talmor and Berant(2019)}]{talmor2019multiqa}
Alon Talmor and Jonathan Berant. 2019.
\newblock {MultiQA}: An empirical investigation of generalization and transfer
  in reading comprehension.
\newblock \emph{arXiv preprint arXiv:1905.13453}.

\bibitem[{Trischler et~al.(2017)Trischler, Wang, Yuan, Harris, Sordoni,
  Bachman, and Suleman}]{trischler2017newsqa}
Adam Trischler, Tong Wang, Xingdi Yuan, Justin Harris, Alessandro Sordoni,
  Philip Bachman, and Kaheer Suleman. 2017.
\newblock {NewsQA}: A machine comprehension dataset.
\newblock \emph{ACL 2017}, page 191.

\bibitem[{Tsatsaronis et~al.(2015)Tsatsaronis, Balikas, Malakasiotis, Partalas,
  Zschunke, Alvers, Weissenborn, Krithara, Petridis, Polychronopoulos
  et~al.}]{tsatsaronis2015overview}
George Tsatsaronis, Georgios Balikas, Prodromos Malakasiotis, Ioannis Partalas,
  Matthias Zschunke, Michael~R Alvers, Dirk Weissenborn, Anastasia Krithara,
  Sergios Petridis, Dimitris Polychronopoulos, et~al. 2015.
\newblock An overview of the {BIOASQ} large-scale biomedical semantic indexing
  and question answering competition.
\newblock \emph{BMC bioinformatics}, 16(1):138.

\bibitem[{Vaswani et~al.(2017)Vaswani, Shazeer, Parmar, Uszkoreit, Jones,
  Gomez, Kaiser, and Polosukhin}]{vaswani2017attention}
Ashish Vaswani, Noam Shazeer, Niki Parmar, Jakob Uszkoreit, Llion Jones,
  Aidan~N Gomez, {\L}ukasz Kaiser, and Illia Polosukhin. 2017.
\newblock Attention is all you need.
\newblock In \emph{Advances in neural information processing systems}, pages
  5998--6008.

\bibitem[{Wang et~al.(2018)Wang, Singh, Michael, Hill, Levy, and
  Bowman}]{wang2018glue}
Alex Wang, Amanpreet Singh, Julian Michael, Felix Hill, Omer Levy, and Samuel~R
  Bowman. 2018.
\newblock {GLUE}: A multi-task benchmark and analysis platform for natural
  language understanding.
\newblock \emph{EMNLP 2018}, page 353.

\bibitem[{Yang et~al.(2019)Yang, Dai, Yang, Carbonell, Salakhutdinov, and
  Le}]{yang2019xlnet}
Zhilin Yang, Zihang Dai, Yiming Yang, Jaime Carbonell, Ruslan Salakhutdinov,
  and Quoc~V Le. 2019.
\newblock {XLNet}: Generalized autoregressive pretraining for language
  understanding.
\newblock \emph{arXiv preprint arXiv:1906.08237}.

\bibitem[{Yang et~al.(2018)Yang, Qi, Zhang, Bengio, Cohen, Salakhutdinov, and
  Manning}]{yang2018hotpotqa}
Zhilin Yang, Peng Qi, Saizheng Zhang, Yoshua Bengio, William Cohen, Ruslan
  Salakhutdinov, and Christopher~D Manning. 2018.
\newblock {HotpotQA}: A dataset for diverse, explainable multi-hop question
  answering.
\newblock In \emph{Proceedings of the 2018 Conference on Empirical Methods in
  Natural Language Processing}, pages 2369--2380.

\bibitem[{Yogatama et~al.(2019)Yogatama, d'Autume, Connor, Kocisky,
  Chrzanowski, Kong, Lazaridou, Ling, Yu, Dyer et~al.}]{yogatama2019learning}
Dani Yogatama, Cyprien de~Masson d'Autume, Jerome Connor, Tomas Kocisky, Mike
  Chrzanowski, Lingpeng Kong, Angeliki Lazaridou, Wang Ling, Lei Yu, Chris
  Dyer, et~al. 2019.
\newblock Learning and evaluating general linguistic intelligence.
\newblock \emph{arXiv preprint arXiv:1901.11373}.

\bibitem[{Yu et~al.(2018)Yu, Dohan, Luong, Zhao, Chen, Norouzi, and
  Le}]{yu2018qanet}
Adams~Wei Yu, David Dohan, Minh-Thang Luong, Rui Zhao, Kai Chen, Mohammad
  Norouzi, and Quoc~V Le. 2018.
\newblock {QANet}: Combining local convolution with global self-attention for
  reading comprehension.
\newblock \emph{arXiv preprint arXiv:1804.09541}.

\end{thebibliography}

\appendix

\end{document}